\def\year{2021}\relax
\documentclass[letterpaper]{article} 
\usepackage{aaai21}  
\usepackage{times}  
\usepackage{helvet} 
\usepackage{courier}  
\usepackage[hyphens]{url}  
\usepackage{graphicx} 
\usepackage{graphicx}  
\usepackage{natbib}  
\usepackage{caption} 
\usepackage{paralist}
\usepackage[switch]{lineno}

\usepackage{subfig}
\usepackage{floatrow}
\usepackage[export]{adjustbox}

\usepackage{amssymb}
\usepackage{amsmath}

\usepackage[vlined, ruled, boxed,linesnumbered]{algorithm2e}
\SetKwInOut{Input}{Input}
\SetKwInOut{Output}{Output}
\SetKwInOut{Parameters}{Params}
\SetKwBlock{Setup}{Setup}{}
\SetKwFor{Loop}{Loop}{}

\usepackage[utf8]{inputenc}
\usepackage[english]{babel}
\usepackage{amsthm}
\theoremstyle{definition}
\newtheorem{definition}{Definition}
\newtheorem*{definition*}{Definition}
\newtheorem{example}{Example}

\usepackage{multirow}
\usepackage{array}
\usepackage{makecell}
\usepackage{hhline}
\usepackage{booktabs}

\newcommand{\norm}[1]{\left\lVert#1\right\rVert}
\newcommand{\reals}{\mathbb{R}}
\newcommand{\nnreals}{\reals^{\ge 0}}

\begin{document}

\title{Scalable and Safe Multi-Agent Motion Planning with \\ Nonlinear Dynamics and Bounded Disturbances}

\author{
    Jingkai Chen\textsuperscript{\rm 1}, 
    Jiaoyang Li\textsuperscript{\rm 2}\protect\thanks{Jiaoyang Li performed the research during her visit at Monash University.}, 
    Chuchu Fan\textsuperscript{\rm 1}, 
    Brian Williams\textsuperscript{\rm 1}
    \\
}
\affiliations{
    \textsuperscript{\rm 1} Massachusetts Institute of Technology \\
    \textsuperscript{\rm 2} University of Southern California \\
    jkchen@csail.mit.edu, jiaoyanl@usc.edu, chuchu@mit.edu, williams@csail.mit.edu
}
\maketitle


\begin{abstract}
We present a scalable and effective multi-agent safe motion planner that enables a group of agents to move to their desired locations while avoiding collisions with obstacles and other agents, with the presence of rich obstacles, high-dimensional, nonlinear, nonholonomic dynamics, actuation limits, and disturbances. We address this problem by finding a piecewise linear path for each agent such that the actual trajectories following these paths are guaranteed to satisfy the reach-and-avoid requirement. We show that the spatial tracking error of the actual trajectories of the controlled agents can be pre-computed for any qualified path that considers the minimum duration of each path segment due to actuation limits. Using these bounds, we find a collision-free path for each agent by solving Mixed Integer-Linear Programs and coordinate agents by using the priority-based search. We demonstrate our method by benchmarking in 2D and 3D scenarios with ground vehicles and quadrotors, respectively, and show improvements over the solving time and the solution quality compared to two state-of-the-art multi-agent motion planners.
\end{abstract}

\section{Introduction}

Multi-agent motion planning has a wide range of real-world applications, but it is notoriously difficult. Even navigating a single robot from an initial location to a goal location while avoiding collisions with obstacles is terribly challenging with the presence of rich obstacles, high-dimensional state space, nonlinear, nonholonomic dynamics, actuation limits, and disturbances. Not to say that when such complex robotic systems can interfere with each other in a shared environment, the scale of this problem is beyond the capability of most existing methods.



In this paper, we present \underline{S}calable and \underline{S}afe \underline{M}ulti-agent \underline{M}otion (S2M2) planner, a novel multi-agent motion planner that can fast and effectively generate provably safe plans for agent models with high-dimensional nonlinear dynamics and bounded disturbances in continuous time and space. Instead of directly planning dynamically-feasible trajectories, which are extremely computationally expensive, S2M2 exploits a separation-of-concerns approach: We first design piecewise linear (PWL) paths $S_i$ for each dynamical agent $\mathcal{A}_i$ to follow, with the understanding that the agents cannot follow those PWL paths exactly. However, we show that with appropriate tracking controllers, the actual trajectories of each agent under disturbances can converge to $S_i$ with guaranteed bounds on the spatial tracking error. More importantly, we show that such error bound can be pre-computed for any qualified PWL path. The secrete sauce behind the high efficiency of S2M2 is that we are able to formulate the problem of finding PWL paths for a single agent as a Mixed Integer-Linear Program (MILP), which can be solved efficiently using off-the-shelf MILP solvers.

To avoid inter-agent collisions, we wrap our single-agent motion planner with the priority-based search~\cite{ma2019searching} that explores the space of priority orderings using systematic search. When priorities are specified, lower-priority agents replan their paths while treating higher-priority agents as moving obstacles. Together with the MILP-based path planner and guaranteed tracking controller, S2M2 can efficiently find plans that are provably safe and robust to disturbances during execution. Moreover, by planning paths for multiple agents on a continuous map over continuous time, our method finds high-quality solutions with low flowtime (i.e., makespan sum of all single-agent plans).

\begin{figure}[t]
    \vspace{4pt}
    \centering
    \includegraphics[width=0.99\columnwidth]{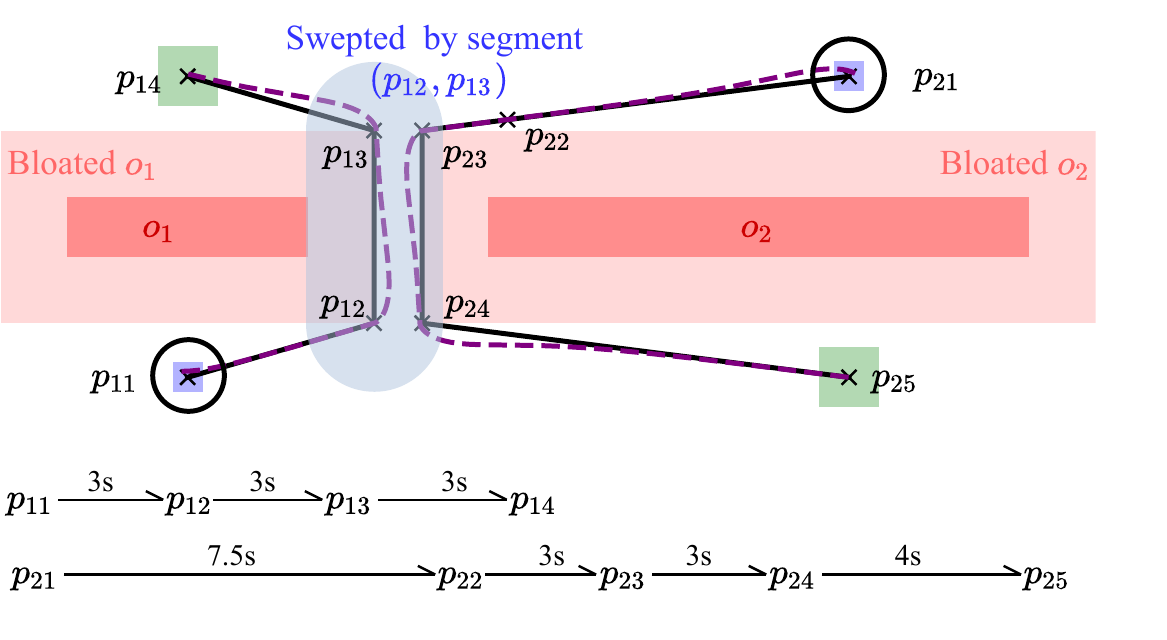}
    \caption{\footnotesize Example of two disc-shaped vehicles making u-turns. The initial and goal position sets are in blue and green, respectively. The obstacles are in red. The bloated obstacles are in light red. The paths of agents $\mathcal{A}_1$ and $\mathcal{A}_2$ are shown in solid black lines, and examples of their actual trajectories are shown in dashed purple lines. The swept area of agent $\mathcal{A}_1$ during its second segment is in light blue.}
    \label{fig:uturn}
\end{figure}

Consider an example of two disc-shaped vehicles making u-turns, as shown in Figure~\ref{fig:uturn}. Given the partially known initial locations and bounded disturbances, we first compute the spatial error bound of tracking a path segment, which is a line, and then consider this error together with the agent shape to obtain the possible swept area of the path segment. The swept area of agent $\mathcal{A}_1$ during its second segment $(p_{12}, p_{13})$ is shown in light blue. We also consider the minimum duration for each segment such that the agent has enough time to adjust its position after turning, which is $3$ seconds in our example. When our single-agent motion planner plans such paths for the agents, the obstacles are bloated with respect to the spatial tracking error and the agent shape, as shown in light red. 
The planner also constrains each segment to respect the minimum segment duration. As we can see, when passing the corridor, both agents slow down and use this $3$ seconds to adjust their trajectories. After passing the corridor, agent $\mathcal{A}_2$ uses its full speed to approach its goal location. When potential collisions are detected, some agents are assigned higher priorities and thus are treated as moving obstacles for others. In our example, agent $\mathcal{A}_1$ is assigned higher priority and thus passes the corridor before agent $\mathcal{A}_2$.

We compare the performance of S2M2 with ECBS-CT \cite{cohen2019optimal} and MAPF/C+POST \cite{honig2018trajectory} on 2D and 3D scenarios with ground vehicles and quadrotors, respectively. Results show that S2M2 outperforms both planners in terms of solution qualities with $15\sim70\%$ reduction for most instances. Moreover, S2M2 requires several magnitudes less time on pre-processing and also shows competitive runtime performance compared to the other two planners. 


\section{Related Work}\label{section:related}

Many works have approached the Multi-agent Motion Planning (MAMP) problem from the perspectives of AI or robotics. These works can be generally divided into two categories with discrete and continuous settings, respectively. In the discrete setting, time and space are discretized into time steps and grids, respectively. Each agent occupies exactly one grid and can only move to adjacent grids at each time step. This problem is known as Multi-agent Path Finding (MAPF)~\cite{SternSoCS19}.
Researchers in the past years have made substantial progress in finding high-quality solutions to various scenarios with hundreds of agents and high congestion
as described in surveys \cite{MaWOMAPF16,felner2017search}. 
Prioritized planning is a popular and widely-used class of MAPF algorithms that coordinate multiple agents by specifying priorities among agents and forcing lower-priority agents to avoid collisions with higher-priority agents by treating their paths as dynamic obstacles \cite{velagapudi2010decentralized,vcap2015complete}. The most recent prioritized planning algorithm Priority-based Search (PBS) \cite{ma2019searching} systematically explores the space of priority orderings, which can lead to near-optimal solutions and possibly scale to a thousand agents. Our method adapts PBS to a continuous setting and thus can efficiently coordinate a large number of agents.

While the classical MAPF assumes synchronized and discretized time, zero-volume shapes, constant velocities, and rectilinear movements, several notable attempts have been made towards closing the gap between the classical MAPF and MAMP using more realistic models. This includes considering the continuous timeline, different-size agent shapes, kinematic constraints, robustness, and any-angle moving directions \cite{walker2018extended,li2019multi,cohen2019optimal,AndreychukYAS19,MaAAAI17,AtzmonAI20,YakovlevA17}. To obtain robust, executable, and high-quality solutions, our method also supports these features.

In the continuous setting, sampling-based motion planners are often used \cite{le2018cooperative,honig2018trajectory}. They first generate a probabilistic roadmap and then apply MAPF algorithms to it. These MAPF solutions are either used to guide the motion tree expansion \cite{le2018cooperative} or post-process to valid continuous trajectories \cite{honig2018trajectory}. Similar to our approach, these algorithms can handle high-order, nonlinear dynamics, and arbitrary complex geometries. Some optimization-based planners tend to solve a large optimization problem in which the decision variables define the trajectories for all agents \cite{augugliaro2012generation,mellinger2012mixed}, which are only demonstrated on small agent teams. 
Our motion planner also uses optimization problems for generating trajectories. However, we only coordinate different agents on demand. To find a feasible plan for each agent, we focus on finding a PWL path satisfying certain duration and boundary requirements, which are encoded as MILPs and solved efficiently.

As a safety guarantee is important to successful execution, safe motion planning is receiving more attention recently. Several approaches are studied in a reference tracking framework, which uses the idea of bounding tracking errors through pre-computation based reachability analysis \cite{herbert2017fastrack,singh2017robust,vaskov2019towards,fan2020fast,majumdar2017funnel}. Other safe motion planners employ barrier functions \cite{barry2012safety,agrawal2017discrete} or use robust model predictive control with chance constraints \cite{blackmore2011chance,jasour2019sequential,yu2013tube}. Some of these works have been extended to the multi-robot setting \cite{wang2016safety,panagou2013multi,srinivasan2018control,desai2019soter,abs-1811-09914,richards2004decentralized}. 

\section{Preliminaries and Problem Statement}\label{section:formulation}

For an $n$-dimensional vector $x \in \mathbb{R}^n$, $x^{(i)}$ is its $i^\text{th}$ entry, $\norm{x}$ is its Euclidean norm, and $B_\epsilon(x) \equiv \{y \in \mathbb{R}^n \|\ \norm{y-x} \leq \epsilon\}$ is the $\epsilon$-ball centered at $x$ with $\epsilon>0$. Given a matrix $H \in \mathbb{R}^{n \times m}$ and a vector $b \in \mathbb{R}^n$, an $(H,b)$-polytope $\texttt{Poly}(H, b)$ is the set $\{x \in \mathbb{R}^m \|\ Hx \leq b\}$. Each row of the polytope defines a halfspace $H^{(i)}x \leq b^{(i)}$, and each face is defined by $H^{(i)}x = b^{(i)}$. $\texttt{dP}(H)$ is the number of its faces.

\begin{definition}[Agent Model]
An agent model $\mathcal{A}_i = \langle  \mathcal{X}_i, \mathcal{U}_i, \mathcal{D}_i, f_i, \mathcal{P}_i\rangle$ is defined by its state space $\mathcal{X}_i \in \mathbb{R}^n$, input space $\mathcal{U}_i \subseteq \mathbb{R}^m$, disturbance space $\mathcal{D}_i \in \mathbb{R}^{l}$, dynamic function $f_i: \mathcal{X}_i \times \mathcal{U}_i \times \mathcal{D}_i \rightarrow \mathcal{X}_i$, and geometric shape $\mathcal{P}_i: \mathcal{X}_i \rightarrow 2^{\mathbb{R}^\delta}$.\footnote{A state is usually made up of positions, orientations, and velocities while an input refers to the input of the controller, such as accelerations and steering rates.}
\end{definition}

The semantics of agent dynamics are defined by trajectories, which describe the evolution of states over time. An input trajectory $u$ over duration $T$ is a continuous function $u$: $[0, T] \rightarrow \mathcal{U}_i$, which maps each time $t \in [0, T]$ to a control signal $u(t) \in \mathcal{U}_i$. Similarly, a disturbance trajectory $d$ over duration $T$ is a continuous function $d$: $[0, T] \rightarrow \mathcal{D}_i$. Given an input trajectory $u$ over $\mathcal{U}_i$, a disturbance trajectory $d$ over $\mathcal{D}_i$, and an initial state $x_{0} \in \mathcal{X}_i$, its state trajectory $\xi_i$ satisfies $\xi_i(x_0, u, d, 0) = x_{0}$ and for all $t > 0$,
\[\small
    \dot \xi_i (x_0, u, d, t) = f_i(\xi_i(x_0, u, d, t), u(t), d(t)).
\]

\begin{example}\label{example:model}
Consider a nonholonomic differential two-wheeled vehicle \cite{rodriguez2014trajectory} as an example. Its state $\xi_i(t)$ consists of three components: $p(t) = [p_x(t), p_y(t)]^T$ is the Cartesian coordinate of the center of inertia, $\theta(t)$ is the angular orientation, and $v(t)$ is the linear velocity. 
We also consider the bounded disturbances $d_x$ on $p_x$, $d_y$ on $p_y$, and $d_\theta$ on $\theta$.
The dynamic function $f_i$ consists of five components: $\dot p_x(t) = v(t) \cos \theta + d_x(t)$, $\dot p_y(t) = v(t) \sin \theta + d_y(t)$, $\dot v(t) = u_1(t) - k v(t)$, $\dot \omega(t) = u_2(t) - k \omega(t)$, and $\dot \theta(t) = \omega(t) + d_\theta(t)$,
where $k$ is a constant, and $u_1(t)$ and $u_2(t)$ are control force and torque as inputs, which can be used to compute the torques for the left and right wheels.
\end{example}

\begin{definition}[MAMP]
A multi-agent motion planning (MAMP) problem is defined by
$\langle \mathcal{W}, O, \mathcal{A},\mathcal{X}^\texttt{init}, \mathcal{X}^\texttt{goal} \rangle \nonumber$, where workspace $\mathcal{W} \subseteq \mathbb{R}^\delta$ is a bounding box in $\mathbb{R}^\delta$; $\delta=2$ for ground vehicles and $\delta = 3$ for aerial and underwater vehicles, and $\xi_i(t) \downarrow \mathcal{W}$ is the projection of $\xi_i(t)$ to the workspace; obstacles $O = \{o_i\}_i \subseteq \mathcal{W}$ are polytopes in $\mathcal{W}$; $\mathcal{A} = \{\mathcal{A}_1,..,\mathcal{A}_N\}$ is a set of agent models; $\mathcal{X}^{\texttt{init}}_i \subseteq \mathcal{X}_i $ and $\mathcal{X}^{\texttt{goal}}_i \subseteq \mathcal{X}_i$ are the initial set and the goal set of  $\mathcal{A}_i$.
The planning problem is to find inputs  $(u_1,..,u_N)$ for every $(x^\texttt{init}_1,..,x^\texttt{init}_N) \in \mathcal{X}^{\texttt{init}}_1  \times \cdots \mathcal{X}^{\texttt{init}}_N$ and every disturbance trajectories $(d_1,..,d_N)$ such that the state trajectories $(\xi_1,..,\xi_N)$ satisfy the
reach-and-avoid requirement: 
\begin{enumerate}
    \item (Dynamics) $\forall \mathcal{A}_i \in \mathcal{A}$, $\xi_i(t) \equiv \xi_i(x^\texttt{init}_i,u_i,d_i,t)$;
    \item (Reach goal set) $\exists t \geq 0$, $\forall \mathcal{A}_i \in \mathcal{A}$, $\xi_i(t) \in \mathcal{X}_{i}^\texttt{goal}$;
    \item (Avoid obstacles) $\forall t \geq 0$, $\forall \mathcal{A}_i \in \mathcal{A}$,  $\mathcal{P}_i(\xi_i(t)) \in \mathcal{W}$ and $\mathcal{P}_i(\xi_i(t)) \cap O = \emptyset$.
    \item (Avoid inter-agent collisions) $\forall t \geq 0$, $\forall \mathcal{A}_i, \mathcal{A}_j \in \mathcal{A}$ and $i \not = j$,   $\mathcal{P}_i(\xi_i(t)) \cap \mathcal{P}_j(\xi_j(t)) = \emptyset$
\end{enumerate}
\end{definition}

In this work, we will solve the MAMP problem by finding a  piecewise linear (PWL) path for each agent. The PWL paths for the agents will be sufficiently far away from the obstacles and from each other so that agents' tracking controllers can drive them to follow their PWL paths to reach their goals without collisions. Here we define PWL paths, tracking controllers, and reachability envelopes of each agent with a given tracking controller.

\begin{definition}[Piecewise Linear Path]
A PWL path $S_i$ in the workspace $\mathcal{W}$ for an agent $\mathcal{A}_i$ is a function $S_i: \nnreals \rightarrow \mathcal{W}$ that maps a time point to a position $S_i(t) \in \mathcal{W}$, which can be constructed from a sequence of time-stamped waypoints $S_i= \texttt{Path} ( \{(t_k, p_k)\}_k)$ such that $S_i(t) = p_{k-1} + \frac{p_k - p_{k-1}}{t_k - t_{k-1}} (t - t_{k-1})$ for $t\in[t_{k-1},t_{k}]$. $(t_k, p_k) \in \nnreals \times \mathcal{W}$ is called the $k^\text{th}$ waypoint of path $S_i$, and $S_i(t)$ when $t\in[t_{k-1},t_{k}]$ is called the $k^\text{th}$ segment of $S_i$ and denoted by $S^{(k)}_i$. 
\end{definition}

\begin{definition}[Decentralized Tracking Controller]
A tracking controller for an agent $\mathcal{A}_i$ is a (state feedback) function $g_i: \mathcal{X}_i \times \mathcal{W} \rightarrow \mathcal{U}_i$. At any time $t$, a tracking controller takes in  a current state of the system $x \in \mathcal{X}_i$ and a desired position $S_i(t) \in \mathcal{W}$, and gives an input $g_i(x, S_i(t)) \in \mathcal{U}_i$ for $\mathcal{A}_i$.
\end{definition}

Fix a tracking controller $g_i$ and a PWL path $S_i$ for an agent $\mathcal{A}_i$, the resulting closed-loop controlled agent becomes an {\em autonomous system}. We use  $\xi^{g_i}_i(x_0, S_i, d_i, t)=\xi_i(x_0, g_i(x_0, S_i(t)), d_i, t)$ to represent the trajectory of the controlled agent $\mathcal{A}_i$ starting from $x_0$ with disturbance $d_i$. The {\em reachablity envelope} of a controlled agent is a set of states around the PWL path that contains all possible actual trajectories of the controlled agent, defined as follows.
\begin{definition}[Reachability Envelope]
    Given an agent model $\mathcal{A}_i = \langle  \mathcal{X}_i, \mathcal{U}_i, \mathcal{D}_i, f_i, \mathcal{P}_i\rangle$, an initial set $\mathcal{X}^\texttt{init}_i \subseteq \mathcal{X}_i$, a PWL path $S_i$, and a tracking controller $g_i$, the reachablity envelope  at time $t$ is 
    $\texttt{Reach}_{\mathcal{A}_i}(\mathcal{X}^\texttt{init}_i,S_i, g_i,  D_i,  t) = 
    \{ \xi^{g_i}_i(x_0,  S_i, d_i, t) \in \mathcal{X}_i \,|\, \exists x_0 \in \mathcal{X}^\texttt{init}_i, \exists d_i: \nnreals \rightarrow \mathcal{D}_i \}$.
\end{definition}

\begin{example}\label{example:controller}
Let a $S(t) = [p^*_x(t), p^*_y(t)]^{T}$ be a PWL path of waypoint sequence $\{(t_k, p_k)\}_k$. From~\cite{rodriguez2014trajectory}, a valid tracking trajectory for Example~\ref{example:model} can be constructed as 
\[\small
\begin{bmatrix}
u_1\\
u_2
\end{bmatrix}=
\begin{bmatrix}
\cos \theta & \sin \theta \\
-\sin \theta/L & \cos \theta/L 
\end{bmatrix}
\begin{bmatrix}
u_1^\prime + v \omega \sin \theta + L \omega^2 \cos \theta \\
u_2^\prime - v \omega \cos \theta + L \omega^2 \sin \theta
\end{bmatrix}
,
\]
where $L$ is a positive constant and $u_1^\prime, u_2^\prime$ are computed as
\[\small
\begin{bmatrix}
u_1^\prime\\
u_2^\prime
\end{bmatrix}=
\begin{bmatrix}
v^* \cos \theta^* - L \sin \theta^* \omega^*\\
v^* \sin \theta^* + L \cos \theta^* \omega^*
\end{bmatrix} +
G\frac{z^{2p-1}}
{1+\|z\|^{2p-1}},
\]
where $G$ is a positive constant, $p$ is a positive integer, {\small$z = \begin{bmatrix}
(p_x - p_x^*) + L (\cos \theta - \cos \theta^*)\\
(p_y - p_y^*) + L (\sin \theta - \sin \theta^*)
\end{bmatrix}$} and $\forall t \in [t_{k-1},t_k], v^*(t) = \frac{\|p_k - p_{k-1}\|}{t_k - t_{k-1}}, \theta^*(t) = \mbox{atan2}(p_{k}-p_{k-1})$.
The dashed purple lines in Figure~\ref{fig:uturn} illustrate such two trajectories of the closed-loop agents.
\end{example}

\section{Approach}\label{section:approach}
\begin{figure}[t]
\centering
\includegraphics[width=0.99\columnwidth]{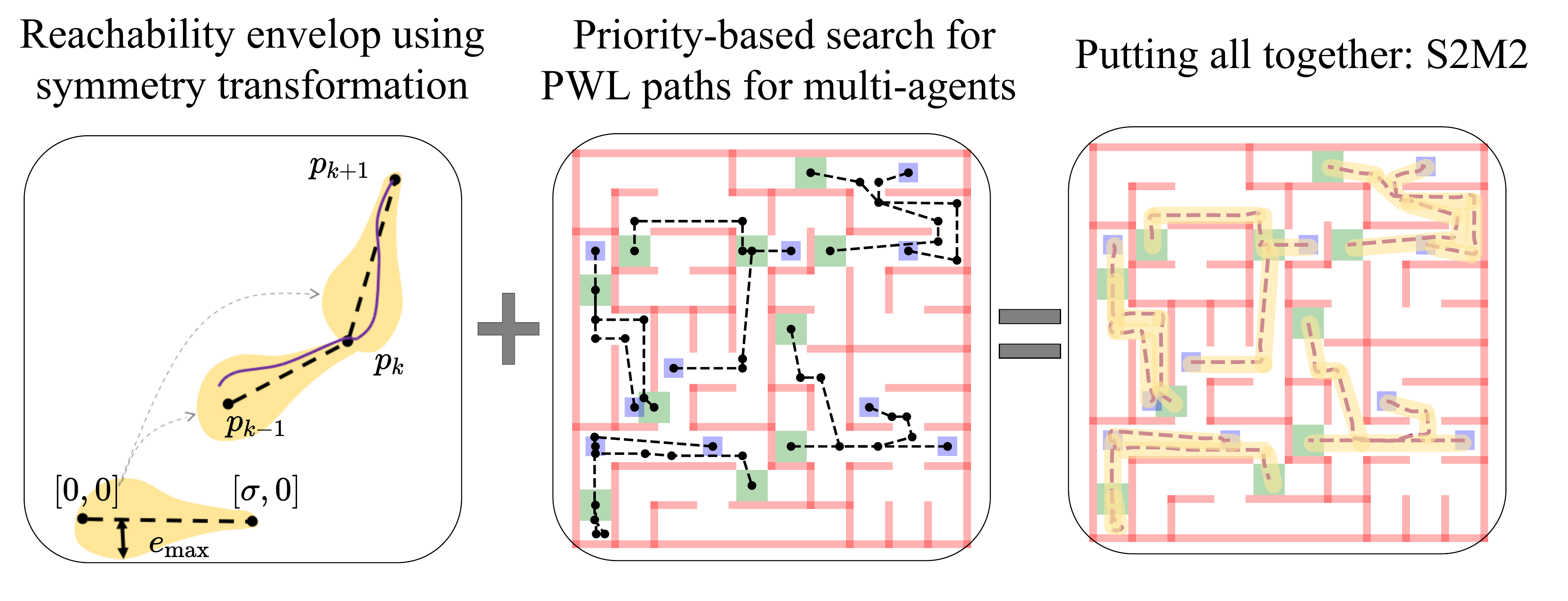}
\caption{\footnotesize Schematic illustration of the approach.}
\label{fig:diagram}

\end{figure}

Figure~\ref{fig:diagram} gives an overview of our approach S2M2, consisting of three key modules: 
\begin{inparaenum}
    \item[(a).] (Figure~\ref{fig:diagram} left): Given a tracking controller $g_i$ for each agent $\mathcal{A}_i$, pre-compute reachability envelopes for any PWL path $S_i$ using symmetry transformations and cashed reachability envelopes, to get two key parameters: (1) an upper bound of the spatial tracking error between the actual trajectory $\xi_i^{g_i}(x_0, S_i, d_i, t) \downarrow \mathcal{W}$ and $S_i$, and (2) the minimum duration of each path segment $S_i^{(k)}$ such that the spatial tracking error bound is always valid (Section \ref{section:approach:reachability}). 
    \item[(b).]  (Figure~\ref{fig:diagram} middle): Given two parameters from (a), the safe motion planning problem for each agent is reduced to finding a PWL path that is sufficiently far from the obstacles and other agents, which is further encoded as a MILP problem (Section~\ref{section:approach:milp}).
    \item[(c).] (Figure~\ref{fig:diagram} middle): To coordinate multiple agents, employ priority-based search to avoid inter-robot collisions, in which some agents treat other agents as moving obstacles and replan their paths optimally (Section~\ref{section:approach:pbs}). 
\end{inparaenum}
Putting them all together (Figure~\ref{fig:diagram} right), S2M2 finds a PWL path for each agent in the multi-agent system, so the closed-loop agents driven by tracking controllers can move along the PWL paths to 
safely achieve the reach-and-avoid requirement. 




\subsection{Computing Reachability Envelopes}\label{section:approach:reachability}

Beyond the fact that a PWL path can be computed efficiently through solving a MILP problem (see Section~\ref{section:approach:milp}), another key idea behind the use of PWL paths is that the reachability envelopes can be pre-computed independent of the concrete values of the PWL paths' waypoints. In this way, we pre-compute the following two parameters of these envelopes, which are used as constraints in finding the PWL paths: (1) an upper-bound of the distance between the actual trajectory projected to the workspace and the reference PWL path; and (2) the minimum duration bound for each path segment to guarantee such distance bound.

Various methods can be used to pre-compute reachability envelopes, including Contraction Metrics~\cite{singh2017robust,tsukamoto2020neural}, Lyapunov functions~\cite{fan2020fast}, Funnels~\cite{majumdar2017funnel}, and Hamilton-Jacobi analysis~\cite{herbert2017fastrack}.
In this section, we take an alternative approach called symmetry transformation. Symmetry transformations on dynamical systems are defined as the ability to compute new trajectories (reachable states) of the same dynamics by applying symmetry operators (e.g., translation and rotation) on existing trajectories (reachable states) \cite{russo2011symmetries}. We show that under mild assumptions, using symmetry transformation, the reachability envelope of an agent  following a PWL path can be constructed from a finite number of reachability envelopes. These envelopes are of the same agent following a single straight line along the x-axis defined by waypoints $(0,[0,0]^T)$ and $(T, [\sigma, 0]^T)$. Then, without knowing the waypoints of a PWL path, we can pre-compute these envelopes around the x-axis and extract the spatial error bound and the minimum segment duration from them.

For the rest of this section, we abuse the notation and use $\texttt{Reach}_{\mathcal{A}_i}^{g_i}(\mathcal{X}^\texttt{init}_{i,k}, S_{i}^{(k)}, t)$ to denote the reachability envelope of agent $\mathcal{A}_i$ following path segment $ S_{i}^{(k)}$ from initial set $\mathcal{X}^\texttt{init}_{i,k}$ at time $t_{k-1}$. The initial set is defined recursively as $\mathcal{X}^\texttt{init}_{i,k}$ $=\texttt{Reach}_{\mathcal{A}_i}^{g_i}(\mathcal{X}^\texttt{init}_{i,(k-1)}, S_{i}^{(k-1)}, t_{k-1})$.
By using a symmetry operator $\gamma$ constructed by translation and rotation, the reachability envelopes of segment $ S_i^{(k)}$ can be constructed from that of segment  $S^{x\text{-axis}}_i$ with the same length and time duration, which is as follows:
\begin{equation*}\scriptsize
    \texttt{Reach}_{\mathcal{A}_i}^{g_i}(\mathcal{X}^k_i, S_i^{(k)},t) = 
    \gamma\left(  \texttt{Reach}_{\mathcal{A}_i}^{g_i} \left(\gamma^{-1}(\mathcal{X}^k_i), S^{x\text{-axis}}_i, t)\right)  \right),
\end{equation*}
where for a set $\mathcal{Y}\subseteq \mathcal{X}$,  $\gamma(\mathcal{Y}) = \{\gamma(x) \ | \ x \in \mathcal{Y}\}$ and $\gamma^{-1}(\mathcal{Y}) = \{ y \ |\ \exists x \in \mathcal{Y}, \gamma(y) = x\}$. 

Since the translation and rotation transformations preserve vector lengths, $\gamma \left(\texttt{Reach}(\cdot,\cdot,\cdot)\right)$ has the same size as  $\texttt{Reach}(\cdot,\cdot,\cdot)$.
Thus, as long as (1)  we pre-compute the reachability envelope that follows a path $S^{x\text{-axis}}_i$ from $\mathcal{X}^\texttt{init}_i$ with duration $T$ and length $\sigma$ for sufficiently long $T$ and all $\sigma \in [v_{\min}T, v_{\max}T ]$; and (2) $\gamma^{-1}(\mathcal{X}^k_i) \subseteq \mathcal{X}^\texttt{init}_i$ for all $k=\{1,..,K\}$ ($K$ is the number of path segments in $S_i$), we can always construct the reachability envelope of following PWL paths using the symmetry operator $\gamma$ on these pre-computed envelopes. These envelopes can be computed using any nonlinear reachability toolbox \cite{chen2013flow,Althoff2018b,fan2017d}. The reachability envelope construction for aerial vehicles or underwater vehicles is similar. 

To understand how far the reference path $S_i$ needs to be away from the obstacles, we define the maximum spatial tracking error $e_{i, \max}$ as follows:
\begin{equation*}\small
\begin{aligned}
&e_{i,\max} = \text{argmin}_{e}  \forall \sigma \in [v_{\min}T, v_{\max}T ], \forall t \in [0,T], \\ &B_{e}(S^{x\text{-axis}}_i(t)) \supseteq 
\texttt{Reach}_{\mathcal{A}_i}^{g_i} \left(\mathcal{X}^\texttt{init}_i, S^{x\text{-axis}}_i,  t\right) \downarrow \mathcal{W},
\end{aligned}
\end{equation*}
where $S^{x\text{-axis}}_i = \texttt{Path}(\{(0,[0,0]^T), (T, [\sigma, 0]^T) \})$.
We implement this equation by first discretizing $[v_{\min}T, v_{\max}T]$ with $\Delta$ and calculating the error bound $e$ for each reachability envelopes with length $\sigma_j = (v_{\min}T+j\Delta)$. Then, we choose the maximum value over all these errors as $e_{i,\texttt{max}}$.

To guarantee that, for any segment $S_i^{(k)}$, the trajectories can converge close enough to it before switching to the next segment, the tracking controller needs to be applied for a sufficiently long time. So we identify the minimum segment duration $T_{i,\min} $ such that $t_k - t_{k-1} > T_{i,\min}$ guarantees $\gamma^{-1}(\mathcal{X}^k_i) \subseteq \mathcal{X}^\texttt{init}_i$, for $k=1, \dots, K$.  Similarly, for the last line segment $S_i^{(K)}$, 
we also find the minimum duration $T_{i,\min}^\prime < t_{K} - t_{K-1} $ such that the final reachable states of the trajectory following path $ S_i^{(K)}$ have enough time to be sufficiently small to fit into the goal set $\mathcal{X}^\texttt{goal}_i$. 

Notice is that for agents with relatively larger actuation limits compared to their velocity ranges, the obtained $e_{i,\text{max}}$ and $T_{i,\min}^\prime$ will be smaller, and our planner will generate paths that are closer to obstacles and make turns more often to minimize the plan makespan. It is also of our users' choices to specify smaller accelerations when computing reachability envelopes, which leads to smoother trajectories but may result in over-conservative plans or even failures. In addition, we can restrict the velocity ranges to have smaller $e_{i,\text{max}}$ and $T_{i,\min}^\prime$, which mitigates the problem of being over-conservative but may lead to plans with longer makespans.

\subsection{Finding Paths for Single Agents}\label{section:approach:milp}

In this section, we describe the method for finding a PWL path for a single agent $\mathcal{A}_i$ with the presence of static obstacles and moving obstacles. To obtain a valid path, we solve a MILP problem. The decision variables of this MILP are $(p_{0}, p_{1}, ..,p_{K})$ with domain $\mathcal{W}$ and $(t_{0}, t_{1}, .., t_{K})$ with domain $\nnreals$, which represent the waypoint positions and their time stamps, respectively. The objective is to minimize the makespan $t_{K}$. We constrain the initial position to be at the center of the initial set $S^\texttt{init}_i$ and the initial time to be $0$: 
    $(p_0 = \texttt{Center}(S^\texttt{init}_i \downarrow \mathcal{W})) \land ( t_0 = 0 )$.

In the rest of this section, we introduce the other sets of the constraints that ensure the instantiated trajectory of the obtained path is valid with respect to the system dynamics, spatial tracking error $e_{i,\texttt{max}}$, minimum segment duration $T_{i,\texttt{min}}$, and minimum duration for the last segment $T_{i,\texttt{min}}^\prime$.

\subsubsection{Time-Position Constraints}
We first add constraints over the duration $(t_k - t_{k-1})$ and the spatial difference $(p_k - p_{k-1})$ for each segment to make sure its velocity $\frac{p_k - p_{k-1}}{t_k - t_{k-1}}$ respects the velocity bound. 
Let $v_\text{min}$ and $v_\text{max}$ be the minimum and maximum allowable velocities of the agent model, respectively. Then, the feasible velocity set is $B_\texttt{vmax}(0) / B_\texttt{vmin}(0)$. We further under-approximate $B_\texttt{vmax}(0)$ to polytope $\texttt{Poly}(H_\texttt{vmax}, b_\texttt{vmax})$, and over-approximate $B_\texttt{vmin}(0)$ to polytope $\texttt{Poly}(H_\texttt{vmin}, b_\texttt{vmin})$. The constraints to ensure each segment to satisfy such dynamic relations are:
\begin{equation*}\small 
    \begin{aligned}
     (\lor_{j=1}^{\texttt{dP}(H_\texttt{vmin})} H_\texttt{vmin}^{(j)}(p_k - p_{k-1}) \geq   b_\texttt{vmin}^{(j)}(t_k - t_{k-1})) \\
    \land (H_\texttt{vmax}(p_k - p_{k-1}) \leq   b_\texttt{vmax}(t_k - t_{k-1}))\
    \forall k = 1,2,..,K
    \end{aligned}
\end{equation*}

We handle disjunctive linear constraints $(\lor_{j=1}^{\texttt{dP}(H)} H^{(j)} x \leq b^{(j)}$ by using the big-M method. We define a $\texttt{dP}(H)$-vector of binary variables $\alpha$, and $\alpha^{(j)} = 1$ if and only if $H^{(j)} x \leq b^{(j)}$ holds for $x$. Let $M$ be a very large positive number, then the constraints is encoded as 
         $(\land_{j=1}^{\texttt{dP}(H)} H^{(j)} x + M(1 - \alpha^{(j)}) \leq b^{(j)}) \land (\sum_{j=1}^{\texttt{dP}(H)} \alpha^{(j)} \geq 1)$.

\subsubsection{Reach-and-Avoid Constraints}

As the actual tracking trajectories deviate from the PWL paths due to inertia, actuation limits, disturbances, and uncertain initial position, we should consider this deviation when encoding constraints related to obstacles and goals.  We have shown that the error between the actual trajectory and the PWL path can be bounded within $e_{i, \text{max}}$ for each agent $\mathcal{A}_i$. In addition to the position deviation, we should consider the agent shape, and the swept area should not intersect with obstacles as well. Then, we know all the possible swept area at position $p$ is $R = p \oplus B_{l}(0)$, where $l = e_{i, \text{max}} + r_i$, and $r_i$ is the radius of the ball containing agent $\mathcal{A}_i$. For obstacle $o = \texttt{Poly}(H_o,b_o) \in O$, the bloated obstacle with respect to $R$ is $\texttt{Poly}(H_{o}, b_o + \norm{H_o}l)$. As long as the path is away from these bloated obstacles, the actual tracking trajectories are guaranteed to be collision-free.


To constrain the segments to be away from an obstacle, which is an polytope, we force the end points of every segment to be at least on one face of this polytope, which is a sufficient condition of being collision-free. The constraint is then as follows: $\forall k = 1,2,..,K, \forall o \in O$,
\begin{equation*}\small
  \begin{aligned}
  \lor_{j=1}^{\texttt{dP}(H_{o})} ((H_{o}^{(j)}p_{k-1} > b_{o}+ \norm{H_o}l)
  \land (H_{o}^{(j)}p_k > b_{o}+ \norm{H_o}l)).
  \end{aligned}
\end{equation*}

For the moving obstacle $o \in O'$ defined by its occupied region $\texttt{Poly}(H_{o}, b_{o})$ and the occupying duration $[lb_o, ub_o]$, we require the agent to either avoid colliding with $o$ or moving through this region out of the duration $[lb_o, ub_o]$:
\begin{equation*}\small
  \begin{aligned}
   \lor_{j=1}^{\texttt{dP}(H_{o})} ((H_{o}p_{k-1} > b_{o}+ \norm{H_o}l) 
  \land (H_{o}p_k > b_{o}+ \norm{H_o}l))\\
    \lor (t_{k-1} < lb_{o})
    \lor (t_k > ub_{o}), \ 
    \forall k = 1,2,..,K, \forall o \in O'.
  \end{aligned}
\end{equation*}

To make sure the spatial error is small enough before switching to the next segment and indeed bounded by $e_{i,\max}$, we add a constraint to force the duration of nonzero-duration segments to be at least $T_{i,\texttt{min}}$ time:
\begin{equation*}\small
    \begin{aligned}
    (t_k - t_{k-1} = 0) \lor
    (t_k - t_{k-1} \geq T_{i,\texttt{min}}) \ \forall k = 1,2,..,K.
    \end{aligned}
\end{equation*}

We also require the agent to be at the goal set $S^\texttt{goal}_i$ at time $t_K$, and the last segment should be at least $T_{i,\texttt{min}}^\prime$ such that the agent has enough time to fit in: 
{\small
     $(p_K = \texttt{Center}(S^\texttt{goal}_i \downarrow \mathcal{W})) \land (t_K - t_{K-1} \geq T_{i,\texttt{min}}^\prime)$.
}

In our MILP encoding with $K$ segments in a $\delta$-dimension workspace, we have $(K+1)(\delta+1)$ or $(K+1)(\delta+1)$ continuous decision variables to represent the states in 2D or 3D state space, respectively. The number of linear constraints and Boolean variables increases linearly in the product of segment number $K$ and the maximum face number of the polytopes that represent the velocity set and obstacles.

\subsection{Coordinating Multiple Agents}\label{section:approach:pbs}

We deploy Priority-based Search (PBS)~\cite{ma2019searching} to coordinate agents and avoid inter-agent collisions. PBS is an efficient two-level search algorithm designed for solving MAPF near-optimally.
When it detects a collision between two agents, it constrains one of the agents to have a lower priority than the other and replans its path by treating the paths of the higher-priority agents as dynamic obstacles. 
As we coordinate the agent trajectories over the continuous timeline and continuous space, which is nontrivial or inefficient to summarize collisions as conflicts, PBS is a more natural candidate to effectively coordinate agents than conflict-based algorithms such as CBS~\cite{sharon2015conflict}. Though PBS does not offer completeness or optimality guarantees, it can explore all the possible priority orderings in theory and find a high-quality solution in few iterations in practice. 

Formally, we coordinate agents and resolve collisions by exploring a priority tree (PT). 
We start with the root PT node, which contains a set of individually optimal paths, not necessarily collision-free, and an empty priority ordering. 
We explore the PT in a depth-first manner, breaking ties in favor of the node with smaller flowtime. During expansion, we identify the pair of agents $\mathcal{A}_i$ and $\mathcal{A}_j$ with the earliest collision and generate two child PT nodes that inherit the priority ordering of the current PT node plus an additional ordered pairs $(j \prec i)$ and $(i \prec j)$, respectively. For each child PT node, we pick the lower-priority agent, i.e., $\mathcal{A}_i$ or $\mathcal{A}_j$, and replan an individually optimal path for it by treating all agents that have higher priorities than it as moving obstacles.
This procedure is terminated when we find a PT node whose paths are collision-free.

Compared to the original PBS in \cite{ma2019searching}, we make two modifications: (1) the single-agent planner in S2M2 uses our MILP encoding since our sub-problem is to find a sequence of time-stamped waypoints in a continuous map over a continuous timeline rather than a graph over discrete time steps; (2) when a new priority is added to resolve a collision, we lazily update the paths by replanning for only the lower-priority agent involved in this collision instead of all the lower-priority agents that have collisions, which shows better scalability in our practical experiments.

\vspace{-4pt}
\begin{figure}[ht]
\centering
\includegraphics[width=0.8\columnwidth]{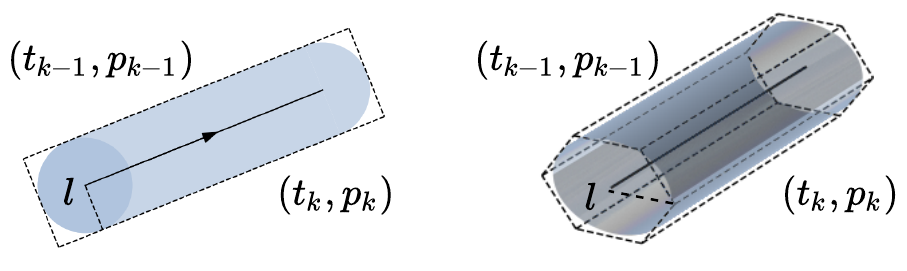}
\caption{\footnotesize Examples of moving obstacles in 2D and 3D workspaces. All the possible swept area is in blue, and is over approximated to polytopes as shown with dashed line boundaries.}
\label{fig:moving}
\end{figure}
\vspace{-5pt}

During replanning, some agents are treated as moving obstacles for other agents. Here we introduce our method for encoding the possible swept area of a path segment as a moving obstacle. Consider a segment from $(t_{k-1}, p_{k-1})$ to $(t_k, p_k)$. As we know all the swept area at position $p$ is $R = p \oplus B_{l}(0)$, we can calculate its moving obstacle as $(t_{k-1}, t_k, \texttt{Poly}(H_k, b_k)$), where $\texttt{Poly}(H_k, b_k)$ is a polygon containing all the possible swept area during duration $(t_{k-1}, t_k)$. Thus, if other agents do not swept $\texttt{Poly}(H_k, b_k)$ during $(t_{k-1}, t_k)$, their paths are guaranteed to be collision-free. The central axis of this moving obstacle is in the direction $\mbox{atan2}(p_{k} - p_{k-1})$ with length $2l + \norm{p_k - p_{k-1}}$. In a 2D workspace, the cross section of this tube is a line that is vertical to the central axis, and its width is $2l$. In a 3D workspace, the cross section is a circle with radius $l$. We further over approximate this circle as a polygon such as a octagon. Figure~\ref{fig:moving} shows examples in 2D and 3D workspace.

\section{Experimental Results}\label{section:results}

We present experimental results on both 2D and 3D environments with ground vehicles and quadrotor models, respectively.  We used DryVR \cite{fan2017d} to generate reachability envelopes and Gurobi 9.0.1 \cite{gurobi2020gurobi} as the MILP solver. We compared S2M2 with the state-of-the-art 2D planner ECBS-CT \cite{cohen2019optimal} and 3D planner MAPF/C+POST \cite{honig2018trajectory}. All experiments were run on a 3.40GHZ Intel Core i7-6700 CPU with 36GB RAM with a runtime limit of $100$s. We repeated each experiment 25 times for each agent number using randomly generated initial and goal locations for the agents. We report the average runtime, success rates (i.e., the percentages of solved instances within the runtime limit), and flowtime (i.e., makespan sum of all single-agent plans) for each agent number on each map.

\vspace{-4pt}

\begin{figure}[ht]
  
    \centering
    \subfloat[Arena.]{\includegraphics[width=0.2\columnwidth]{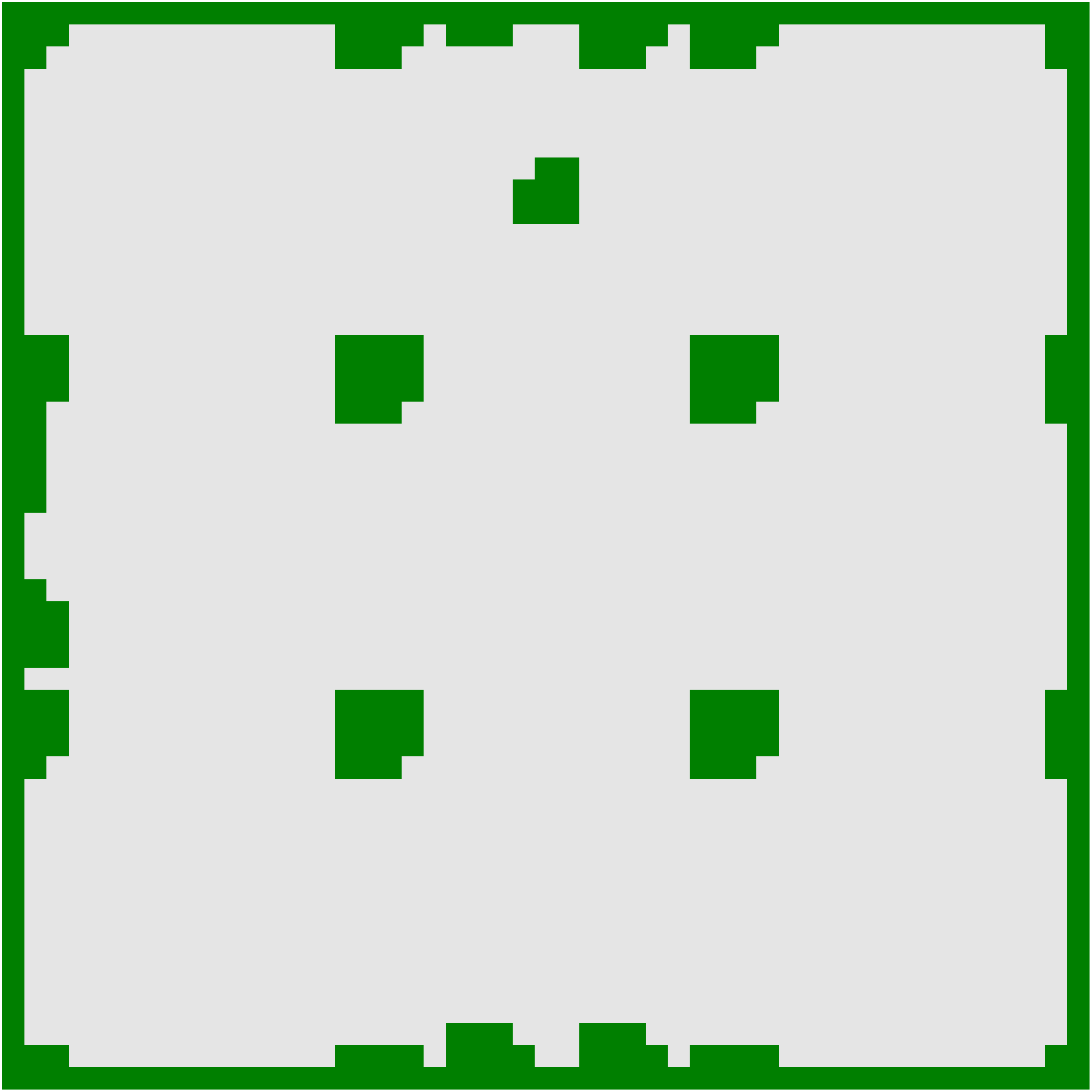}}
    \,\,\,\,\,\,
    \subfloat[Den502d.]{\includegraphics[width=0.2\columnwidth]{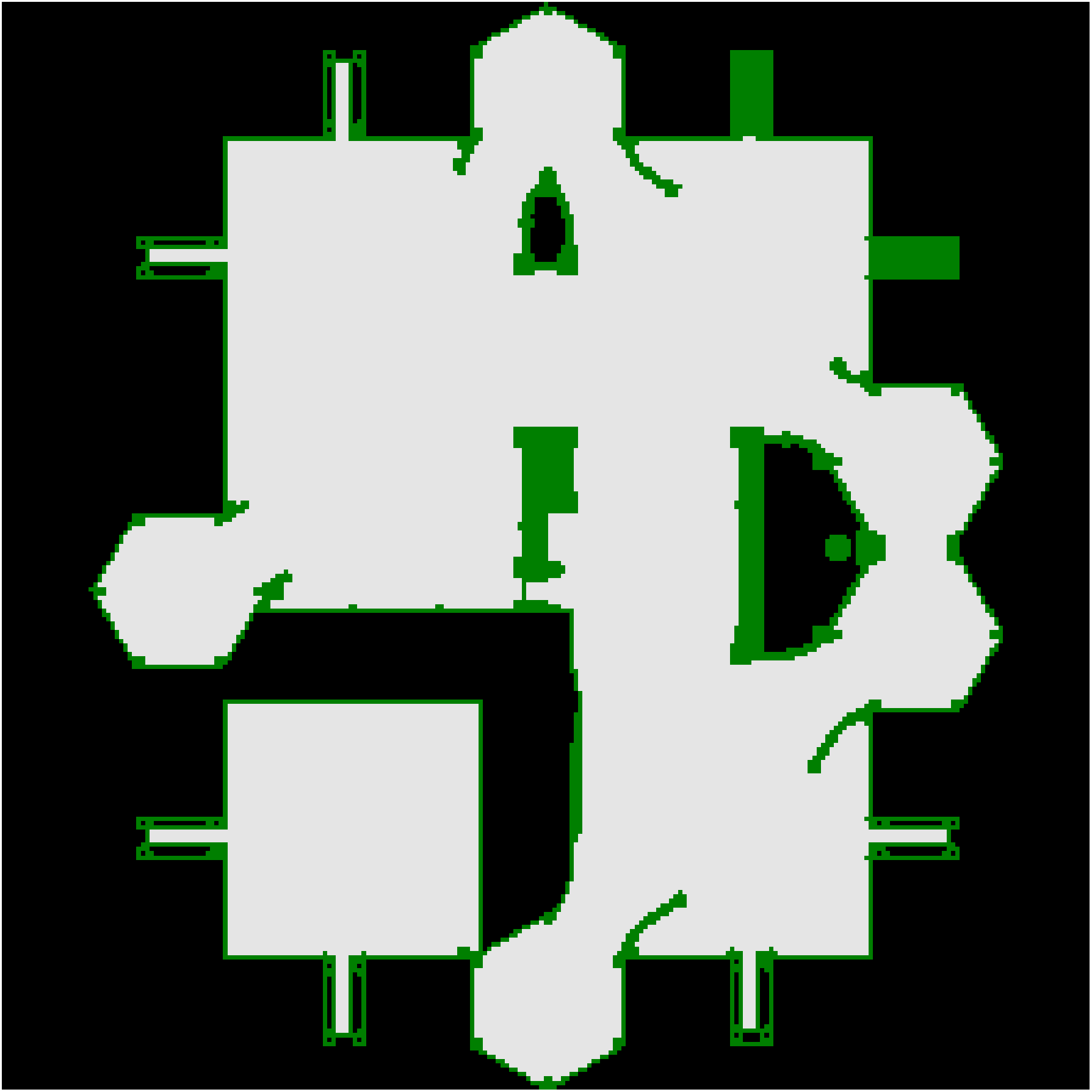}}
    \,\,\,\,\,\,
    \subfloat[Wall.]{\includegraphics[width=0.28\columnwidth]{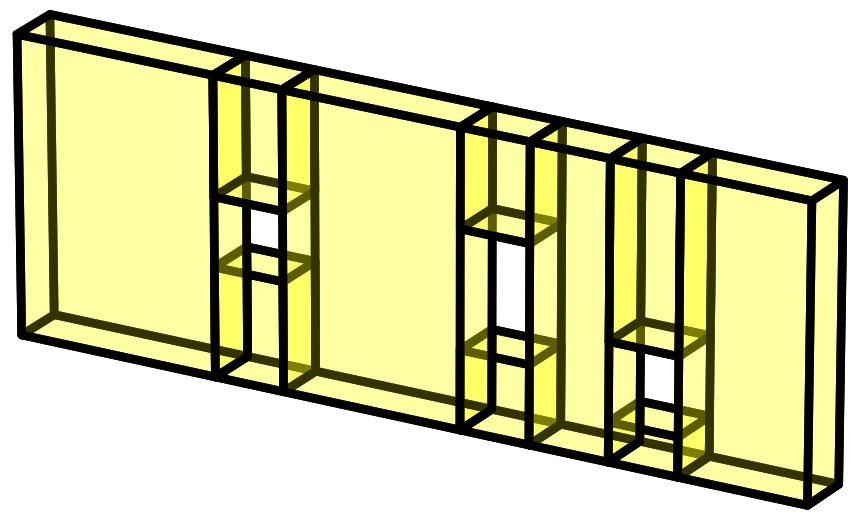}}
    \caption{\footnotesize Maps and scenarios.}
    \label{fig:map}

\end{figure}

\vspace{-15pt}

\subsubsection{2D Experiment}
\begin{figure}[t]
    \centering
    \subfloat[Runtimes for Arena.]{\includegraphics[width=0.48\columnwidth]{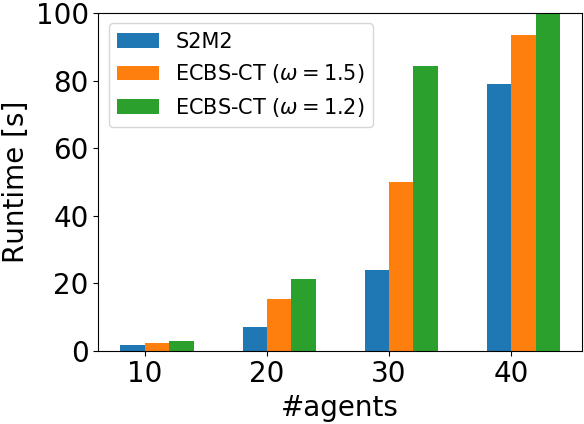}}
    \,
    \subfloat[Success rates for Arena.]{\includegraphics[width=0.48\columnwidth]{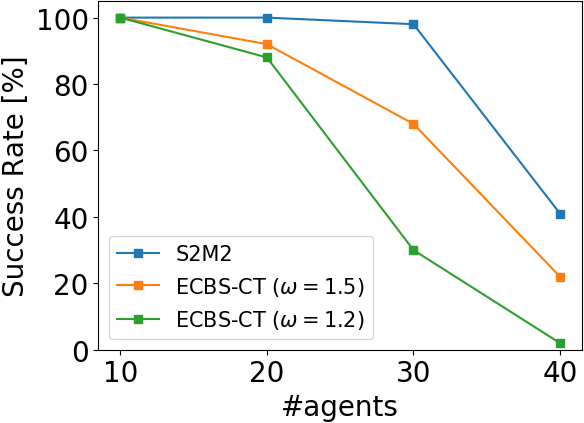}}
    \,
    \subfloat[Runtimes for Den502d.]{\includegraphics[width=0.48\columnwidth]{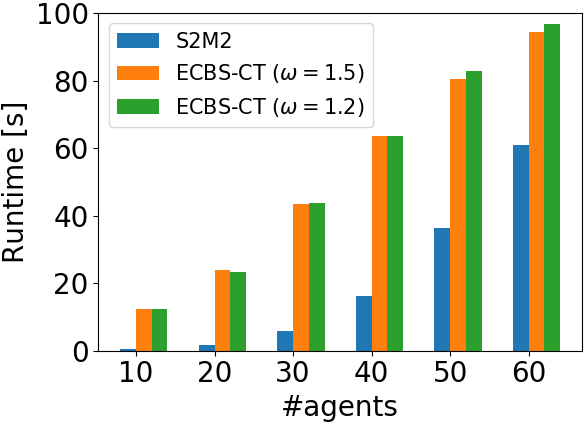}}
    \,
    \subfloat[Success rates for Den502d.]{\includegraphics[width=0.48\columnwidth]{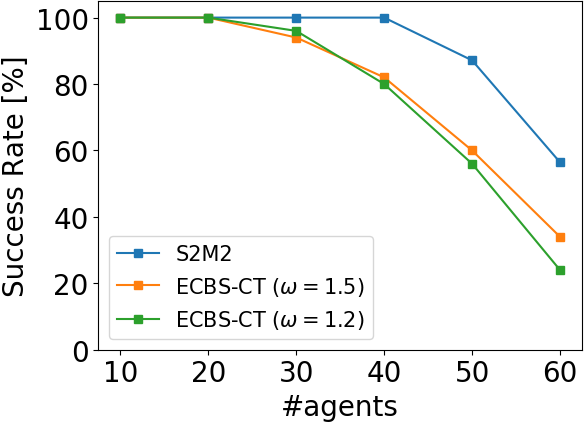}}
    \,
    \subfloat[Runtimes for Wall.]{\includegraphics[width=0.48\columnwidth]{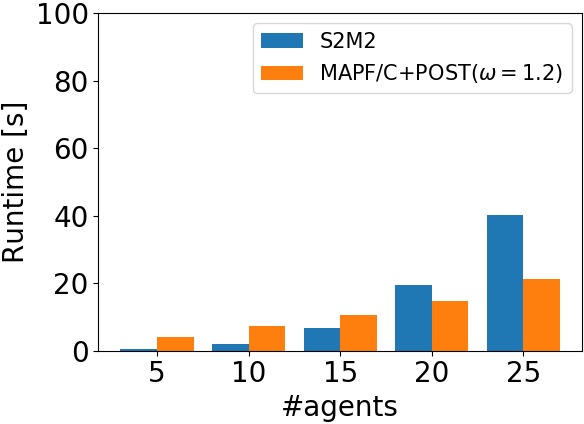}}\
    \,
    \subfloat[Success rates for Wall.]{\includegraphics[width=0.48\columnwidth]{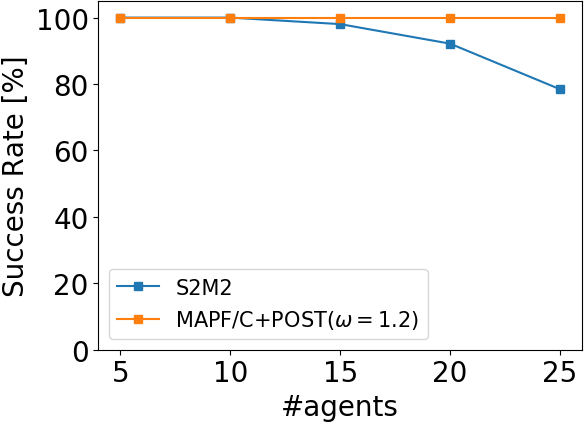}}
    \caption{\footnotesize Runtimes and success rates.}
    \label{fig:time}
\end{figure}

We compare S2M2 against ECBS-CT on two benchmark maps from the Grid-Based Path Planning Competition (GPPC)\footnote{GPPC: https://movingai.com/GPPC}, namely Arena (49 $\times$ 49) and Den502d (251 $\times$ 211). 
We assume a disc-shaped agent with a radius of $0.5$. While agent dynamics for ECBS-CT is approximated with sixteen discrete orientations and five primitives, which is taken from the Search-based Planning Laboratory (SBPL)\footnote{
SBPL: http://sbpl.net}, S2M2 considers a vehicle model with continuous, nonholonomic, nonlinear dynamics from \cite{rodriguez2014trajectory} with additional bounded disturbances.  The cost multiplier for the motion primitive model is set to $1$, which means no preferred action is specified. We set the maximum velocity in both planners to be $1$. We set the focal weight $\omega$ (i.e., suboptimality ratio) for ECBS-CT to $1.2$ and $1.5$. The average runtimes and success rates of our method and ECBS-CT on these two maps are given in Figure~\ref{fig:time}(a)-(d). The corresponding solution qualities are given in Table~\ref{table:cost}. Note that the pre-computation time is discussed separately and not added to the average runtime.

First of all, we observe that S2M2 is several magnitudes faster than ECBS-CT in terms of the pre-processing time. While S2M2 takes $0.33s$ to pre-compute the spatial error bound and the minimum duration for all agents, the time for ECBS-CT to pre-compute the heuristic is $0.27s$ on Arena and $18.18s$ on Den502d for each agent, which makes the total pre-processing time for ECBS-CT very large. For example, for each instance on Den502d with 60 agents, ECBS-CT spends roughly $1,100s$ to compute these heuristics. 

Furthermore, S2M2 outperforms ECBS-CT in terms of both runtimes and success rates, 
as shown in Figure~\ref{fig:time}(a)-(d). While S2M2 halves the runtime of ECBS-CT on Arena for most instances when $\omega = 1.5$, it is roughly one-third of that when $\omega = 1.2$ (Figure~\ref{fig:time}(a)). In Figure~\ref{fig:time}(c), the runtime of ECBS-CT does not change much with different weights on Den502d. While S2M2 takes half the time than ECBS-CT for most Den502d instances, S2M2 is one magnitude faster when the agent number is less than $40$. As shown in Figure~\ref{fig:time}(b)(d), the success rates of ECBS-CT drop much faster than S2M2. When ECBS-CT fails half the instances, S2M2 still solves more than $90\%$ of them. We do not include the case when $\omega = 1$ (i.e., search for optimal solutions) since ECBS-CT merely solves instances even for $10$ agents. We also test with larger focal weights, but it does not show significant improvement over runtimes or success rates.

In Table~\ref{table:cost}, we can see S2M2 reduces at least half the flowtime for Arena instances, and this reduction is up to $70\%$ as the agent number increases. This is because S2M2 directly plans high-fidelity models on a continuous map over continuous timeline, which provides more flexibility in avoiding collisions, especially in smaller maps. On the larger map Den502d, we can still observe roughly $15\%$ cost reduction.  

\begin{table}[t]\scriptsize
\centering





{
\begin{tabular}{|c|c||c|c|c|}
\hline
 & \#agents & S2M2  & \makecell{ECBS-CT \\ ($\omega=1.2$)} & \makecell{ECBS-CT \\ ($\omega=1.5$)}  \\ \hhline{|=|=||=|=|=|}
\multirow{5}{*}{\rotatebox[origin=c]{90}{\centering Arena}} 
& 10 & 382.02 & 867.83 & 868.00  \\ \cline{2-5} 
& 20 & 741.20 & 1848.07 & 1948.07 \\ \cline{2-5} 
& 30 & 1062.60 & 2929.95 & 3147.36 \\ \cline{2-5} 
& 40 & 1366.23 & NA & 4554.60 \\ \hhline{|=|=||=|=|=|}
\multirow{5}{*}{\rotatebox[origin=c]{90}{\centering Den502d}}
& 10 & 1292.01 & 1566.78 & 1567.87 \\ \cline{2-5} 
& 20 & 2569.11 & 3135.57 & 3134.96 \\ \cline{2-5} 
& 30 & 3962.72 & 4621.11 & 4620.03 \\ \cline{2-5} 
& 40 & 5517.90 & 6202.30 & 6292.29 \\ \cline{2-5} 
& 50 & 7289.37 & 7798.79 & 7821.35 \\ \cline{2-5} 
& 60 & 8681.73 & 9452.26 & 9480.53 \\ \hline
\end{tabular}}

\vspace{4pt}

{
\begin{tabular}{|c|c||c|c|c|}
\hline
 & \#agents & S2M2  & \makecell{MAPF/C} & \makecell{MAPF/C+POST}  \\ \hhline{|=|=||=|=|=|}
\multirow{5}{*}{\rotatebox[origin=c]{90}{\centering Wall}} 
& 5 & 48.00 & 50.32 & 80.35  \\ \cline{2-5} 
& 10 & 102.77 & 103.08 & 142.27  \\ \cline{2-5} 
& 15 & 162.37 & 152.38 & 200.36 \\ \cline{2-5} 
& 20 & 230.75 & 200.67 & 230.00 \\ \cline{2-5} 
& 25 & 299.29 & 265.09 & 315.20  \\ \hline
\end{tabular}
}
\caption{\footnotesize Solution quality (i.e., average flowtime) for Arena and Den502d (above) and Wall (below).}
\label{table:cost}
\end{table}

\subsubsection{3D Experiment}
We use map Wall ($13 \times 13 \times 5$) from \cite{honig2018trajectory}.  In this scenario, a nano-quadrotor team \cite{preiss2017crazyswarm} starts from one side of the wall with three windows and is asked to fly to the other side. We compare S2M2 with MAPF/C+POST \cite{honig2018trajectory}, which performs a generalized-MAPF algorithm called MAPF/C on a graph with sparse samples and then post-processes the discrete solution to valid continuous trajectories. We set the maximum velocity in both planners to be $1$. We use the same parameters for sampling and post-processing as the original paper. The focal weight of MAPF/C is chosen to be $1.2$. In post-processing, we set the total iterations to $7$ and continuity degree to $4$. The average runtime and success rate of our method and MAPF/C+POST are given in Figure~\ref{fig:time}(e)-(f) while the solution qualities are given in Table~\ref{table:cost}.

While it takes $0.83s$ for S2M2 to pre-compute the error bound and minimum duration, the time to sample roadmaps for MAPF/C+POST is $668.94s$ and annotating conflicts takes $902.00s$, which is $1570.94s$ in total. This pre-processing time is such long because its sampling and annotating procedures, which are critical to the efficiency and solution qualities of MAPF/C+POST, requires computationally expensive distance checking on all pairs of edges. Thus, MAPF/C+POST is sensitive to the map size and only scales to small maps. We also tested its sampling module with the 3D Arena map ($49 \times 49 \times 5$), in which the obstacle height is $5$. We failed to get a reasonably connected roadmap in hours. 


Although MAPF/C+POST is efficient on sparse, well-connected roadmaps, S2M2 can still solve most instances faster when \#agents $<20$. The runtime is less than $1$s for instances with $10$ agents. In Table~\ref{table:cost}, we also observe that S2M2 has up to $40\%$ cost reduction compared to MAPF/C+POST trajectories. Even though the discrete MAPF/C solution is bounded suboptimal, 
the quality of its valid continuous trajectory is not guaranteed.

\section{Conclusions}
We presented S2M2, a fast and effective multi-agent motion planner that generates provably safe plans for agent models with high-dimensional, nonlinear dynamics and bounded disturbances. S2M2 plans piecewise linear paths that satisfy certain safe bounds and coordinates multiple agents using the priority-based search. We show that S2M2 improves both the solving time and the solution quality compared to two state-of-the-art multi-agent motion planners ECBS-CT and MAPF/C+POST. Especially, S2M2 saves much time on pre-processing either for computing heuristics or sampling roadmaps compared to the existing methods.

\subsubsection{Acknowledgement}
This work at Massachusetts Institute of Technology was supported by Kawasaki Heavy Industries, Ltd (KHI) under grant number 030118-00001.

\bibliography{bib}
\end{document}